\documentclass[10pt,twocolumn,letterpaper]{article}

\usepackage{iccv}              

\usepackage[dvipsnames]{xcolor}

\definecolor{iccvblue}{rgb}{0.21,0.49,0.74}
\usepackage[pagebackref,breaklinks,colorlinks,allcolors=iccvblue]{hyperref}

\usepackage{float}
\usepackage{amsmath}
\usepackage{xcolor,colortbl}
\usepackage{comment}
\usepackage{adjustbox}
\usepackage{commath}
\usepackage{amssymb}
\usepackage{xcolor}
\usepackage{pifont}
\usepackage{multirow}
\usepackage{algorithm}
\usepackage{algpseudocode}
\usepackage{hyperref}
\usepackage{soul}

\newcommand{\floor}[1]{\left\lfloor #1 \right\rfloor}

\def\paperID{5755} 
\def\confName{ICCV}
\def\confYear{2025}

\title{BillBoard Splatting (BBSplat): Learnable Textured Primitives \\ for Novel View Synthesis}

\author{
David Svitov\textsuperscript{1,2}
\quad
Pietro Morerio\textsuperscript{2} 
\quad 
Lourdes Agapito\textsuperscript{3} 
\quad
Alessio {Del Bue}\textsuperscript{2}\\\\
\textsuperscript{1}Università degli Studi di Genova,  Genoa, Italy \\
\textsuperscript{2}Istituto Italiano di Tecnologia (IIT), Genoa, Italy \\
\textsuperscript{3}Department of Computer Science, University College London \\
{\tt\small \{david.svitov, pietro.morerio, alessio.delbue\}@iit.it}\quad {\tt\small l.agapito@cs.ucl.ac.uk}
\vspace{-2em} 
}

\begin{document}
\makeatletter
\g@addto@macro\@maketitle{
  \vspace{-0.5em}
  \begin{figure}[H]
  \setlength{\linewidth}{\textwidth}
  \setlength{\hsize}{\textwidth}
  \centering
  \includegraphics[width=12cm]{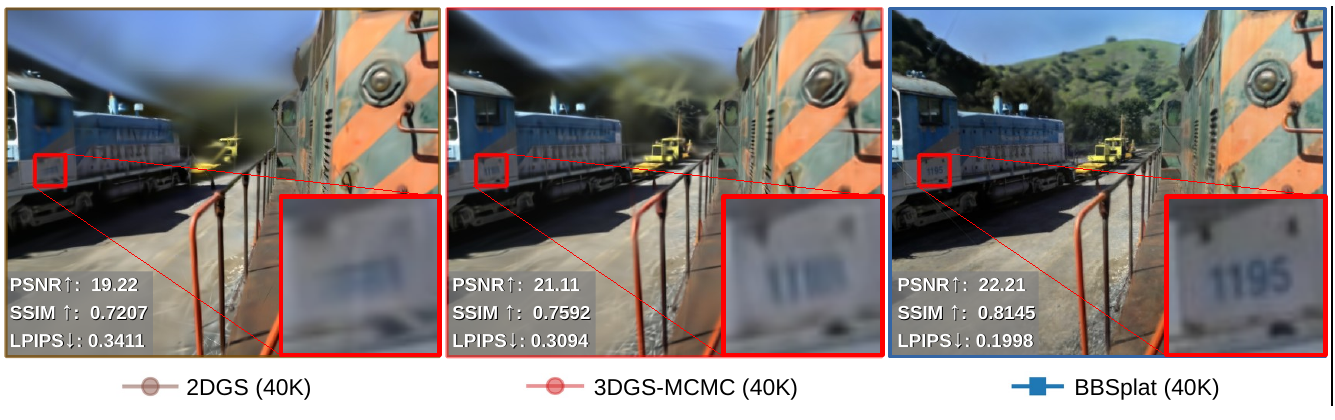}
  \includegraphics[width=4.5cm]{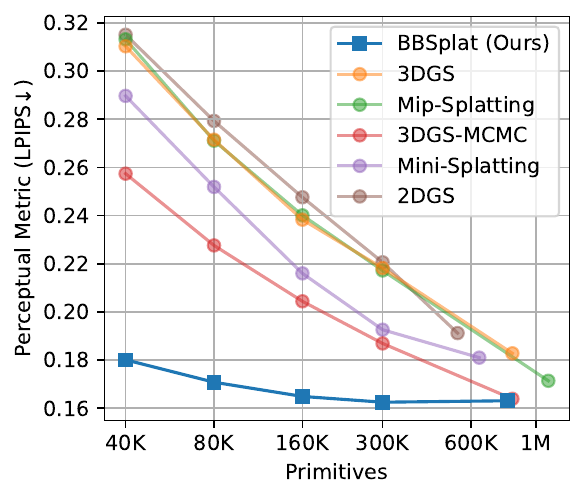}
  \caption{\textbf{BBSplat has a competitive advantage on novel view synthesis (NVS) with less primitives.} Left: BBSplat brings more detailed NVS than 3DGS and 2DGS with equal number of Gaussians, \eg background and planar regions have higher quality as they are better modeled by BBSplat's textured primitives.  
  Right: The plot shows perceptual similarity score (the lower, the better) at a varying number of primitives.   BBSplat show better performance than state-of-the-art methods, with a more evident gap with less primitives.} 
  \label{fig:shapes}
  \end{figure}
}
\makeatother

\maketitle

\begin{abstract}
We present billboard Splatting (BBSplat) - a novel approach for novel view synthesis based on textured geometric primitives. BBSplat represents the scene as a set of optimizable textured planar primitives with learnable RGB textures and alpha-maps to control their shape. BBSplat primitives can be used in any Gaussian Splatting pipeline as drop-in replacements for Gaussians. 
The proposed primitives close the rendering quality gap between 2D and 3D Gaussian Splatting (GS), enabling the accurate extraction of 3D mesh as in the 2DGS framework. Additionally, the explicit nature of planar primitives enables the use of the ray-tracing effects in rasterization.
Our novel regularization term encourages textures to have a sparser structure, enabling an efficient compression that leads to a reduction in the storage space of the model up to $\times17$ times compared to 3DGS. Our experiments show the efficiency of BBSplat on standard datasets of real indoor and outdoor scenes such as Tanks\&Temples, DTU, and Mip-NeRF-360. Namely, we achieve a state-of-the-art PSNR of 29.72 for DTU at Full HD resolution.
Project page: \href{http://david-svitov.github.io/BBSplat_project_page}{david-svitov.github.io/BBSplat\_project\_page}.
\end{abstract}

\section{Introduction}
\label{sec:intro}
Novel view synthesis (NVS) is a crucial technology 
for a variety of applications, including virtual reality, computer gaming, and cinematography. 
Several efforts have been dedicated to deploy methods that are more efficient while providing a better quality of the synthesized images. Specifically, the choice of geometric primitives used to represent the scene plays a key role in defining the advantages and drawbacks of different NVS methods.
Recent breakthroughs on neural representations \cite{Lombardi:2019, Lombardi21, mildenhall2021nerf, barron2021mip, barron2022mipnerf360, mueller2022instant, watsonnovel_iclr2023, chen2023mobilenerf, hu2023tri, chen2022tensorf, yu2021plenoxels} have been sided by recent methods based on Gaussian Splatting \cite{huang20242d, kerbl20233d, kheradmand20243d, Yu2023MipSplatting, fang2024mini, niemeyer2024radsplat, Fu2023COLMAPFree3G, zhang2024fregs}, demonstrating most efficient way for novel view rendering.

Indeed, NeRF-based methods \cite{mildenhall2021nerf, barron2022mipnerf360, barron2021mip, mueller2022instant}
still achieve the best NVS quality for challenging real-world captures by using implicit scene representations such as the weights of an MLP. However, image rendering with a NeRF  is less efficient as it requires repeated MLP inferences to predict colors along camera rays. An alternative to neural rendering, 3D Gaussian Splatting (3DGS) \cite{kerbl20233d} uses faster rendering based on projecting explicit primitives on the screen surface while preserving high-quality NVS. In practice, 3DGS uses Gaussian-distributed radiance around explicit 3D scene points as primitives. 

Recently, 2D Gaussian Splatting (2DGS) \cite{huang20242d} proposed the use of flat Gaussians, oriented in 3D as flat primitives align better with the object surface. 
Since 2D Gaussians are effectively tangent to object surfaces, they allow more accurate surface extraction. Despite proving their efficiency in mesh extraction tasks,  
2D primitives result in a downgrade of rendering metrics compared to 3D primitives such as 3D Gaussians. In this work, we aim to make 2D primitives suitable for high-quality NVS by introducing a new primitive representation, but still allowing reliable mesh extractions.

Our proposed geometric primitives for NVS take inspiration from the classic \textit{billboards} used for extreme 3D model simplification \cite{decoret2003billboard} by replacing mesh during a greedy optimization process. A 3D scene can be efficiently rendered using a ``billboard cloud'' of textured planar primitives with alpha channels. Using billboards, we can efficiently model planar surfaces, such as a painting on a wall or scene background, while dramatically reducing the number of geometric primitives required, up to an order of magnitude with respect to 3DGS/2DGS (check \cref{fig:shapes} for a comparison). 

\begin{figure*}
    \centering
    \includegraphics[width=17cm]{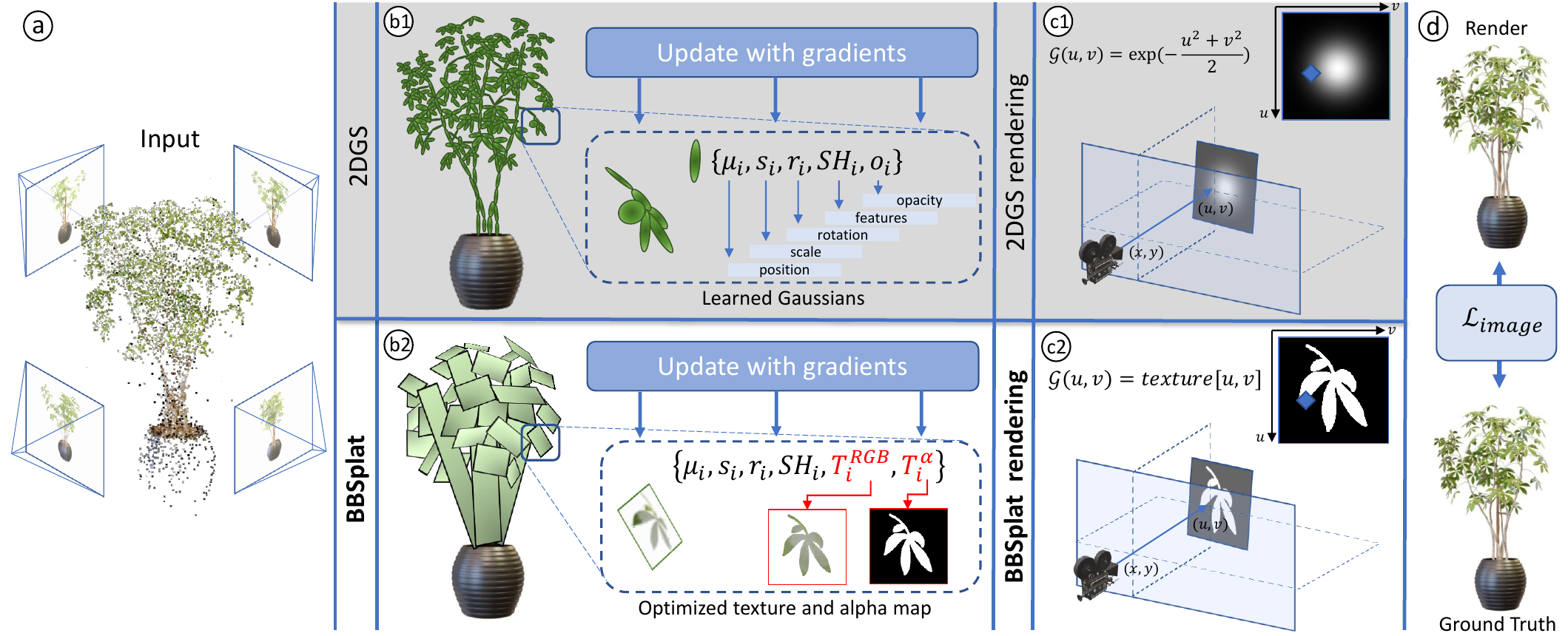}
    \caption{\textbf{Method description.} a) As input we use point cloud and camera positions predicted with COLMAP~\cite{schoenberger2016sfm}. b) Our BBSplat parametrization (b2) extends Gaussian primitives parametrization (b1) with two textures for each point: RGB texture for colors and alpha texture for transparency. c) As defined in 2DGS splatting technique (c1), we find the ray-plane intersection, but instead of calculating Gaussian opacity, we sample color and opacity from the texture (c2). d) To train our 3D scene representation, we utilize only photometric losses.}
    \label{fig:main_scheme}
\end{figure*}

We define \emph{billboards} with 2D Gaussians parameters (rotation, scaling, 3D center location, and spherical harmonics) while also introducing RGB texture and alpha map to control pixel-wise color and shape (\cref{fig:main_scheme}(b2)).
The alpha map defines the billboard silhouette and models the arbitrary shape of primitives. Similarly, the RGB texture stores color for each point of the billboard. In this way, we can use fewer primitives to represent high-frequency details (\cref{fig:shapes}). The key aspect of our approach, BillBoard Splatting (BBSplat), is a method to learn billboard parameters from a set of calibrated images. 

To tackle the challenge of  storing all billboard textures, we compress them by representing each texture as sparse offsets from colors calculated by spherical harmonics \cite{adelson1991plenoptic} and by further quantizing them to 8 bits. Then, we can efficiently utilize dictionary-based compression algorithms~\cite{shannon1948mathematical, fano1949transmission} for quantized textures. 
  
To summarise, our contributions are as follows:
\begin{itemize}
    \item We propose BBSplat with optimizable textured primitives to learn 3D scene representation with photometric losses. The proposed representation surpasses 2DGS and 3DGS in rendering quality, especially for smaller numbers of primitives.
    \item The BBSplat representation allows extracting reliable meshes from a scene with accuracy superior to 3D Gaussian-based methods. 
    \item We developed an algorithm to efficiently represent and store textures for billboards. BBSplat demonstrates significant storage cost reduction.
\end{itemize}

\section{Related work}
\label{sec:related_work}

Reconstruction of 3D scene for novel-view synthesis is a long-standing problem that used to be solved with structure from motion approaches (SfM) \cite{schoenberger2016sfm, zhao2018linear, snavely2008modeling, fuhrmann2014mve, snavely2006photo}. The significant increase in quality was achieved with the appearance of implicit scene representation \cite{Lombardi:2019, mildenhall2021nerf, park2019deepsdf}. 

Nowadays, Neural Radiance Fields (NeRF) \cite{mildenhall2021nerf} is the most commonly used implicit method for NVS. It uses a neural network to predict points' colors and transparency based on points coordinates along the camera ray. 
The biggest drawback of such representation is rendering speed, as it requires multiple inferences of an MLP. 
InstantNGP \cite{mueller2022instant} achieves the most acceleration of NeRF up to 150 FPS by using multi-scale feature grids.

To overcome the rendering speed issue, the 3D Gaussian Splatting (3DGS) \cite{kerbl20233d} method proposed an explicit method for 3D scene representation. Based on the point cloud reconstructed by SfM \cite{schoenberger2016sfm}, they train the color and orientation of ellipsoids with Gaussian distributed transparency. For NVS, these ellipsoids are splatted on the screen surface providing high rendering speed. As a result, 3DGS has become popular for many practical real-time applications such as human avatars \cite{svitov2024haha, lei2024gart, xu2023gaussianheadavatar} or SLAM \cite{Matsuki:Murai:etal:CVPR2024, yugay2023gaussianslam}.

Recent works further improve 3DGS in terms of rendering quality. Thus, in \cite{bulo2024revising} authors revise the densification algorithm to reduce the number of artifacts by using loss-based splitting of Gaussians. In 3DGS-MCMC \cite{kheradmand20243d}, authors use Markov Chain Monte Carlo interpretation of sampling algorithm to more efficiently distribute Gaussians over the scene, even with random initialization. 
RadSplat \cite{niemeyer2024radsplat} uses pre-trained NeRF representation as a prior during 3DGS training to improve rendering quality in challenging scenarios. 
Mip-Splatting \cite{Yu2023MipSplatting} solves the issue of gaps between Gaussian produced due to changes in camera distance by introducing a 3D smoothing filter that constrains Gaussian size.

Some of the recent works also propose to disentangle appearance and geometry for 3DGS. Texture-GS \cite{xu2024texture} introduces an MLP to establish correspondence between 3D Gaussian surface and texture map. In practice this approach allows one to edit the texture in the scene but is limited to synthetic scenes with one object and no background. Another approach \cite{huang2024textured} encodes color and opacity as spherical harmonics on the surface of 3D Gaussian, which allows one to improve reconstruction fidelity. 

2D Gaussian splatting (2DGS) \cite{huang20242d} was proposed to increase the quality of mesh surface reconstruction from such scene representation. Specifically, 2DGS uses ellipses instead of ellipsoids as they align along the object's surface and make mesh extraction operation more straightforward. For rendering of such flat primitives, 2DGS utilizes a ray-splat intersection algorithm \cite{weyrich2007hardware, sigg2006gpu}. While 2DGS increases the accuracy of mesh reconstruction, its fidelity falls shortly behind 3DGS, especially for in-the-wild scenes. In this work, we investigate the ability of flat primitives to be used for high-quality novel-view rendering.

Billboard clouds \cite{decoret2003billboard} propose an extreme 3D model simplification by replacing it with a set of textured planes. This representation is widely used in computer graphics for real-time forest rendering \cite{bao2009billboards, fuhrmann2005extreme, lacewell2006stochastic} and found implementation in such tasks as automatic paper slice-form design \cite{le2013automatic, mccrae2011slices, ruiz2014multi} and level of detail control \cite{holst2007surfel}. In this work, we take inspiration from the billboard clouds technique \cite{decoret2003billboard} and propose arbitrarily shaped textured primitives as an extension of the 2DGS approach. 

\section{Method}
\label{sec:method}

\subsection{Preliminaries: 2D Gaussian Splatting}
\label{sec:preliminaries}

Recently proposed 2D Gaussian splatting (2DGS) \cite{huang20242d} uses 2D flat ellipses (in contrast with 3D ellipsoids for 3DGS) orientated along the object's surface to represent the scene. These ellipses have Gaussian distributed transparency and are parametrized as $\{\mu_i, s_i, r_i, o_i, \textrm{SH}_i\}$ which corresponds to position, 2D scale, quaternion rotation, global opacity, and spherical harmonics of the primitive as in Fig. \ref{fig:main_scheme}(b1). Here spherical harmonics \cite{adelson1991plenoptic, fridovich2022plenoxels} define view-dependent color. In practice, to calculate ellipses orientation, 2DGS uses local tangent planes defined with transformation matrix $H_i$ as:
\begin{equation}
    \mathcal{P}_i(u, v) = H_i(u, v, 1, 1)^T =\begin{bmatrix}
                R_iS_i & \mu_i\\
                0  & 1
              \end{bmatrix} (u, v, 1, 1)^T,
\end{equation}
where matrices $R_i$ and $S_i$ are given by quaternion rotation $r_i$ and scale $s_i$, and $(u, v)$ is the point coordinate on the plane $\mathcal{P}_i$. Then, 2DGS utilizes $(u, v)$ coordinates of points within a plane to calculate 2D Gaussian intensity as follows:
\begin{equation}
    \mathcal{G}(u, v) = exp (-\dfrac{u^2 + v^2}{2})
    \label{eq:2dgs}
\end{equation}

To project Gaussians on the screen, 2DGS finds $(u, v)$ coordinates corresponding to the screen pixel as the intersection of a camera-direction ray with a plane $\mathcal{P}_i$. For efficiency, 2DGS utilizes an explicit ray-splat intersection algorithm proposed by \cite{sigg2006gpu}.

\subsection{Billboard Splatting}
In this work, we propose to further improve NVS efficiency and quality by introducing billboard splatting. These billboard primitives are based on planes, but leverage learnable textures instead of utilizing Gaussian distribution (\cref{eq:2dgs}) to calculate their color and transparency. We inherit 2DGS parametrization for these planes: $\{\mu_i, s_i, r_i, \textrm{SH}_i\}$, where $\mu_i$ is the center of $i$-th plane, $s_i$ is scales along two axis, $r_i$ is rotation quaternion, $\textrm{SH}_i$ is spherical harmonics coefficients. Instead of using global Gaussian opacity $o_i$ we set transparency at each point of a plane as $T_i^\alpha$ texture (\cref{fig:main_scheme}~(b2)). This way, each billboard can have an arbitrary shape. Additionally, we parametrize billboards with $T_i^\textrm{RGB}$ - a RGB texture to control the color of all points of the billboard. 

We use explicit ray-splat intersection algorithm \cite{sigg2006gpu} to find the corresponding plane coordinate $\textbf{u} = (u, v)^T$ for $\textbf{x} = (x, y)^T$ screen coordinate. Specifically, we parametrize the camera ray as an intersection of two 4D homogeneous planes: $\textbf{h}_x = (-1, 0, 0, x)^T$ and $\textbf{h}_y = (0, -1, 0, y)^T$. We then transform these planes to the $uv$-coordinate system of plane $H$ to find intersection point in $(u, v)$ coordinates: 
\begin{equation}
    \textbf{h}_u = (WH)^T\textbf{h}_x \hspace{10 mm} \textbf{h}_v = (WH)^T\textbf{h}_y,
\end{equation}
where $W \in 4 \times 4$ is the transformation matrix from world space to screen space. Then we calculate the $(u, v)$ coordinate of the intersection point as:
\begin{equation}
    \textbf{u}(x) = \dfrac{\textbf{h}_u^2 \textbf{h}_v^4 - \textbf{h}_u^4 \textbf{h}_v^2}{\textbf{h}_u^1 \textbf{h}_v^2 - \textbf{h}_u^2 \textbf{h}_v^1} \hspace{5 mm} \textbf{v}(x) = \dfrac{\textbf{h}_u^4 \textbf{h}_v^1 - \textbf{h}_u^1 \textbf{h}_v^4}{\textbf{h}_u^1 \textbf{h}_v^2 - \textbf{h}_u^2 \textbf{h}_v^1},
\end{equation}
where $\textbf{h}_u^i, \textbf{h}_v^i$ are the $i$-th value of the 4D homogeneous plane parameters.

This way, we get $(u, v)$ coordinates in the $[-1; 1]$ range with $0$ corresponding to the plane center $\mu_i$. After that, we rescale them to the range $[0; S_T]$, where $S_T$ corresponds to texture size in texels. Using rescaled $(u, v)$ coordinates, we sample \textit{color} and \textit{opacity} of the ray-splat intersection point (\cref{fig:main_scheme} (c2)) and accumulate them along the ray as:
\begin{equation}
    c(x) = \sum_{i=1} c_i[\textbf{u}(x)] T_i^\alpha[\textbf{u}(x)] \prod_{j=1}^{i-1} (1 - T_j^\alpha[\textbf{u}(x)]),
\end{equation}

where $c_i$ is the view-dependent color calculated with $\textrm{SH}_i$ and sampled color $T_i^\textrm{RGB}[\textbf{u}(x)]$. To sample textures' values $T[\cdot]$, we utilize \textit{bilinear sampling} \cite{smith1981bilinear}. This way, we take into account neighboring texels and can calculate gradients for billboard position $\mu_i$ with respect to the texture. We adopt PyTorch \cite{NEURIPS2019_9015} CUDA implementation of \textit{bilinear sampling} to sample texture values and calculate gradients. Namely, we redistribute input gradient values between texels that contributed to the sampled value according to their contribution weights (\ie bilinear coefficients). 

\begin{figure}
    \centering
    \includegraphics[width=7cm]{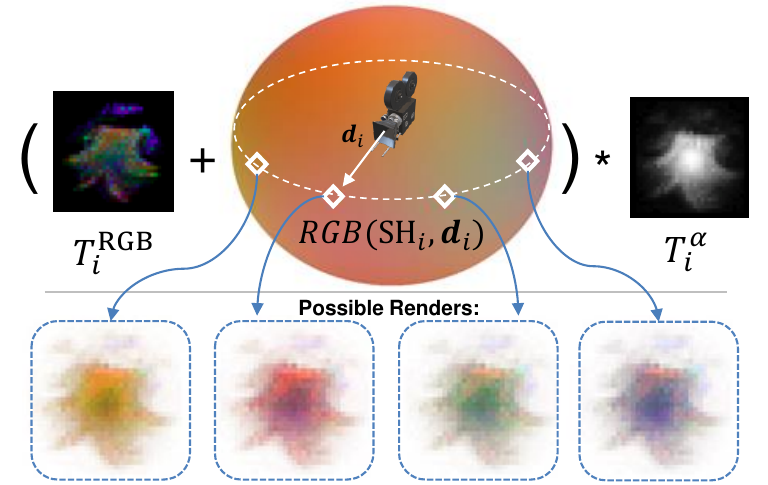}
    \caption{\textbf{Texture rasterization.} During rendering to get final billboard colors we add texture $T^\textrm{RGB}_i$ colors to base color calculated with $\textrm{SH}_i$ and view-direction vector $\textbf{d}_i$. It results in textured billboards handling light effects. Here we showcase four possible renderings for different camera directions.}
    \label{fig:sh}
\end{figure}

To calculate value of view-dependent colors $c_i$ at each uv-point \textbf{u} of $i$-th billboard, we integrate sampled texture colors $T_i^\textrm{RGB}[\textbf{u}]$ as offsets to the colors predicted with $\textrm{SH}_i$ and view-direction vector $\textbf{d}_i$:
\begin{equation}
    c_i[\textbf{u}] = T_i^\textrm{RGB}[\textbf{u}] + RGB(\textrm{SH}_i, \textbf{d}_i).   
    \label{eq:color}
\end{equation}
Here $\textrm{SH}_i$ is a spherical harmonic features \cite{adelson1991plenoptic} for each primitive, and $\textbf{d}_i$ is the view-direction vector from the camera position to billboard center position $\mu_i$. We transform $\textrm{SH}_i$ to RGB color space and define this operation as $RGB(\cdot, \cdot)$. This way, we can handle view-dependent light effects for textured planes (\cref{fig:sh}). Representing the final color as a combination of spherical harmonics and sampled color was initially proposed in \cite{xu2024texture} and helps to disentangle texture and light effects. In our case it also allows $T_i^\textrm{RGB}$ to have a sparse structure, which is more efficient to store as described in \Cref{seq:compression}.


Final parametrization of billboard is as follows: $\{\mu_i, s_i, r_i, \textrm{SH}_i, T_i^\textrm{RGB}, T_i^\alpha\}$, where $\{\mu_i, s_i, r_i, \textrm{SH}_i\}$ follows 2DGS parametrization (\Cref{sec:preliminaries}) and the values $\{T_i^\textrm{RGB}, T_i^\alpha\}$ correspond to our proposed textures for color and opacity.

The use of alpha maps in billboard splitting is crucial for accurate mesh extraction as it allows for detailed scene representation (\cref{fig:blender_mesh} (b)) even for a limited number of billboards. We then render depth maps corresponding to training views and fuse them with truncated signed distance fusion (TSDF) for mesh extraction, using Open3D \cite{zhou2018open3d}.

\subsection{Training}

To train billboard splatting representation of the scene, we initialize them with sparse point cloud obtained using SfM \cite{schoenberger2016sfm}.
Optionally, we add 10K evenly distributed points on the circumscribing sphere using the Fibonacci algorithm \cite{gonzalez2010measurement} to represent the sky and far away objects. We use only photometric losses to fit a scene as in Gaussian splatting pipelines. For textures, we propose regularizations to avoid their overfitting and get a more sparse structure that reduces storage costs. In this section, we describe the losses and regularizations we use.

\textbf{Photometric losses.}
We utilize photometric losses to train our billboard representation. Namely we apply $\mathcal{L}_1$ and structure similarity D-SSIM losses:
\begin{equation}
    \mathcal{L}_\textrm{image} = (1-\lambda_\textrm{SSIM})\mathcal{L}_1 + \lambda_\textrm{SSIM} \mathcal{L}_\textrm{SSIM}.
    \label{eq:photometric}
\end{equation}

Therefore, we need only a set of images to train the BBSplat representation of the scene. 

\textbf{Regularizations.}
Due to the large number of parameters, textures tend to overfit training images, which results in noisy rendering output for novel views. To alleviate this issue, we propose a simple yet efficient regularization in which we push billboards with low impact on the renderings (\ie small or far billboards affecting few pixels) to have Gaussian distributed transparency.

First, we define per-texture visibility weight $w_i$ for all $N$ billboards based on their impact $I_i$ on the rendered image. We define the impact for $i$-th billboard as a sum over alpha-blending values corresponding to its rendered pixels $\mathcal{R}_i$: 

\begin{equation}
\begin{split}
    I_i & = \sum_{x \in \mathcal{R}_i} \left( T_i^{\alpha}[\textbf{u}(x)] \cdot \prod_{j=1}^{K_x} \left( 1 - T_j^{\alpha}[\textbf{u}(x)] \right) \right), \\
    w_i & =  
    \begin{cases}
        \sigma - min(I_i, \sigma), & \text{if } I_i > 0\\
        0,              & \text{otherwise}
    \end{cases}
\end{split}
\end{equation}

Alpha-blending values are calculated similarly to 2DGS and take into account $K_x$ overlapping planes at pixel $x$. We use the maximum impact threshold $\sigma=500$ to regularize only clearly visible billboards. The choice of the threshold is discussed in the \cref{sec:details_params} of Supp. Mat.. By ignoring billboards with $I_i \leq 0$, we do not regularize invisible in the frame primitives, avoiding over-regularization in underrepresented areas. The proposed criterion is more efficient than the one utilizing projected radii as in 2DGS since they only take hitting frustum into account and ignore overlapping.

Finally, we define the regularization term to push RGB textures close to zero and alpha-maps close to a Gaussian distribution $\mathcal{G}$ based on their visibility weight $w_i$:

\begin{equation}
\begin{split}
    \mathcal{L_\textrm{RGB}} & = \dfrac{1}{N} \sum_{i=0}^N w_i \norm{T_i^\textrm{RGB}}, \\
    \mathcal{L_\alpha} & = \dfrac{1}{N} \sum_{i=0}^N w_i \norm{T_i^\alpha - \mathcal{G}}, \\
    \mathcal{L_\textrm{texture}} & = \mathcal{\lambda_\textrm{RGB} L_\textrm{RGB} + \lambda_\alpha L_\alpha}.
\end{split}
    \label{eq:texture_reg}
\end{equation}

With this regularization, the final loss to train BBSplat 3D scene representation is defined as:

\begin{equation}
    \mathcal{L = L_\textrm{image} + L_\textrm{texture}}.
\end{equation}

\subsection{Adaptive density control} \label{sec:BBSplat_MCMC}
Most 3DGS-based methods use adaptive density control to adjust the number of Gaussians. Specifically, they perform splitting, cloning, and pruning operations according to user-set policies. An alternative approach is proposed in \cite{kheradmand20243d}, where they reformulate Gaussian Splatting densification as a Markov Chain Monte Carlo (MCMC) sampling. MCMC introduces noise in the training process and replaces splitting and cloning with a deterministic state transition. 

We also adopt and modify the MCMC sampling technique for BBSplat. In the sampling stage, MCMC clones Gaussians and adjusts their opacity $o_i$ and scale set by covariation matrix $\Sigma_i$ to preserve the rendering output state:

\begin{equation}
o_{1,..,N}^{new} = 1 - \sqrt[N]{1 - o_N^{old}},
\vspace{-1.5em}
\end{equation}

\begin{multline}
    \Sigma_{1,...,N}^{new} = \\
    (o_N^{old})^2 { \left( \sum_{i=1}^N \sum_{k=0}^{i-1} \left(\binom{i-1}{k} \frac{(-1)^k(o_N^{new})^{k+1}}{\sqrt{k+1}} \right)\right)}^{-2} \Sigma_N^{old}.
    \label{eq:scale_adjustment}
\end{multline}

While this adjustment is well fit for Gaussians, we noticed that scale adjustment in \cref{eq:scale_adjustment} is unsuitable for arbitrarily shaped primitives and violates the rendering state preservation. Instead, we adjust only $T_i^\alpha$ to preserve overall transparency along the ray, because it controls transparency in all points on a plane and therefore does not require additional scale modification:

\begin{equation}
T_{1,..,N}^{\alpha} = 1 - \sqrt[N]{1 - T_N^{\alpha}}.
\end{equation}

This makes the adjustment on parameters more straightforward compared to \cref{eq:scale_adjustment}. Finally, where MCMC compares opacity $o_i$ with a predefined threshold $\gamma=5\mathrm{e}{-3}$ to determine ``dead'' Gaussians, we compare the average value of the alpha-map: $\overline{T_i^{\alpha}} < \gamma$.

\subsection{Texture compression}
\label{seq:compression}
Using textured splats leads to new challenges of efficiently storing them. For each point, we use two textures of size $S_T \times S_T$ with one channel for transparency and three channels for color. Considering that textures are stored with 4-byte float point values, we need $(S_T*S_T + S_T*S_T*3) * 4$ bytes of memory to store each texture.

To reduce storage costs, we first apply quantization \cite{jacob2018quantization} of texture parameters. Let us define normalized in $[0; 1]$ range texture values as $\epsilon$. For storing BBSplat, we quantize texture values by rescaling $\hat{\epsilon} = \lfloor \epsilon * 255 \rceil$ where $\lfloor \cdot \rceil$ is the rounding operation. When loading textures, we apply an inverted operation $\epsilon = \hat{\epsilon} / 255$ for dequantization. This way, we can store each texel value $\hat{\epsilon}$ as a single byte, reducing storage costs four times.

Our final color representation described in \cref{eq:color} as a sum of color calculated with SH and texture sampled offsets allows us to additionally reduce storage costs.  
Our regularization term $\mathcal{L_\textrm{texture}}$ proposed in \cref{eq:texture_reg} allows us to get sparser $T^\textrm{RGB}$ values and $T^\alpha$ closer to a Gaussian distribution $\mathcal{G}$. We increase the sparsity of $T^\alpha$ by subtracting the Gaussian pattern $T^\alpha - \mathcal{G}$ and ensure more zero values in the texture. The sparse nature of textures allows us to use efficient dictionary-based compression methods \cite{shannon1948mathematical, fano1949transmission} (\eg~ZIP \cite{zipformat}) to further reduce storage costs. Our texture compression pipeline allows us to reduce memory consumption $\sim 7$ times on average.

\begin{figure*}
    \centering
    \includegraphics[width=17.5cm]{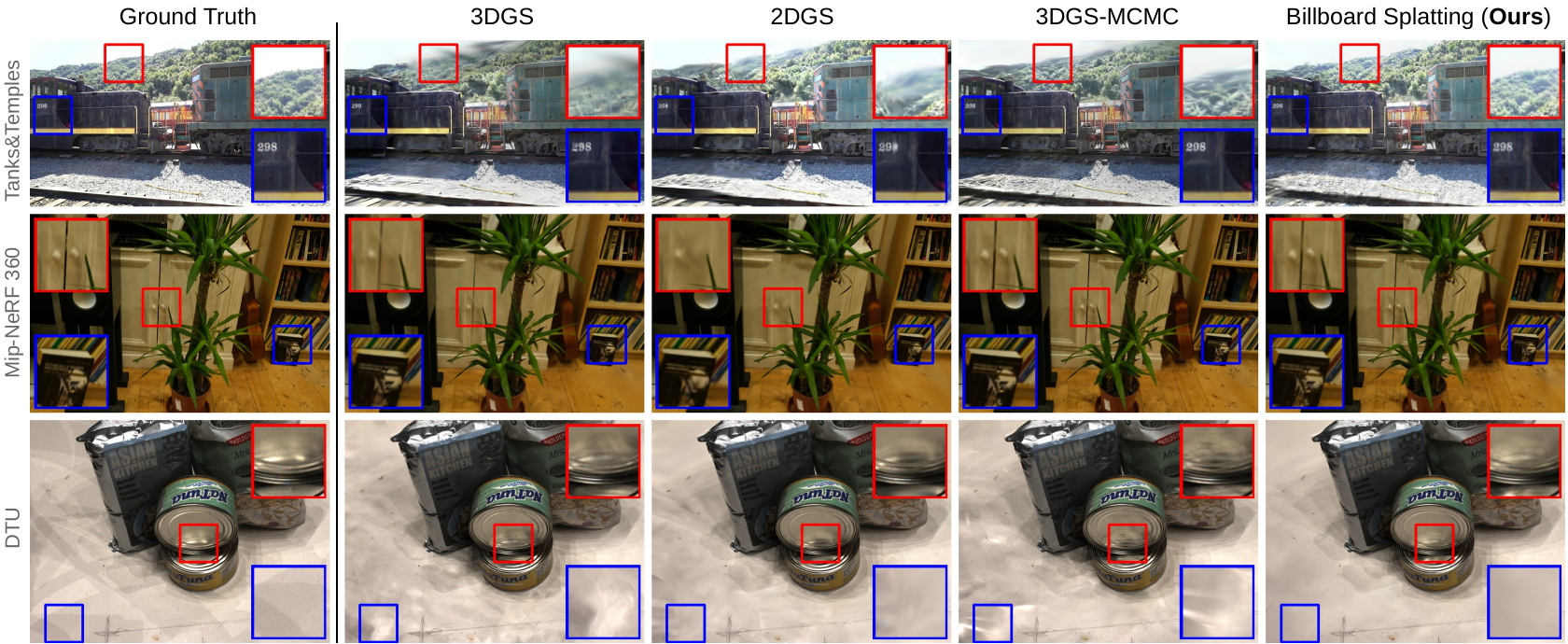}
    \caption{\textbf{Qualitative results.} We provide rendering results of scenes from each dataset: Tanks\&Temples [Train], Mip-NeRF-360 [Room], and DTU [Scan97]. For competitors methods, we use the maximum number of Gaussians recommended by the method. 
    More renderings can be found in the \cref{sec:results_images} of Supp. Mat..}
    \label{fig:results}
\end{figure*}

\section{Experiments}
\label{sec:experiments}

We compared our billboard splatting representation against five state-of-the-art 3D scene representation methods: 3DGS \cite{kerbl20233d}, Mip-Splatting \cite{Yu2023MipSplatting}, 3DGS-MCMC \cite{kheradmand20243d}, Mini-Splatting \cite{fang2024mini}, and 2DGS \cite{huang20242d}. For 2DGS, we conducted two sets of experiments: with and without depth-normal regularizations, as they affect metrics differently for various datasets. In the following section, we provide both quantitative and qualitative comparisons for indoor and outdoor real scenes from standard datasets: Tanks\&Temples \cite{Knapitsch2017}, Mip-NeRF-360 \cite{barron2022mipnerf360}, and DTU \cite{jensen2014large}. Please check supplementary material for additional images, quantitative results, and further details on experimental implementation and evaluation.

\begin{table*}[tb]
\centering
\begin{adjustbox}{width=1\textwidth}
\setlength{\tabcolsep}{2pt}
\begin{tabular}{l|cccccc|cccccc|cccccc}
     & \multicolumn{6}{c|}{DTU}   & \multicolumn{6}{c|}{Tanks\&Temples} & \multicolumn{6}{c}{Mip-NeRF-360} \\ \hline
     & PSNR$\uparrow$  & SSIM$\uparrow$   & LPIPS$\downarrow$ & Storage$\downarrow$ & FPS$\uparrow$ & Points$\downarrow$ & PSNR$\uparrow$  & SSIM$\uparrow$   & LPIPS$\downarrow$ & Storage$\downarrow$ & FPS$\uparrow$ & Points$\downarrow$ & PSNR$\uparrow$  & SSIM$\uparrow$   & LPIPS$\downarrow$  & Storage$\downarrow$ & FPS$\uparrow$ & Points$\downarrow$ \\ \hline
3DGS~\cite{kerbl20233d} & 26.10 & 0.8482 & 0.3452 & 7 MB & 690 ± 77 & 30K & 20.90 & 0.7814 & 0.3130 & 19 MB & 1362 ± 165 & 80K & 25.25 & 0.7800 & 0.3179 & 38 MB & 866 ± 191 & 160K \\ 
Mip-Splatting~\cite{Yu2023MipSplatting} & 25.69 & 0.8433 & 0.3464 & 7 MB & 593 ± 82 & 30K & 20.78 & 0.7801 & 0.3162 & 19 MB & 1054 ± 123 & 80K & 25.15 & 0.7782 & 0.3207 & 38 MB & 672 ± 142 & 160K \\
3DGS-MCMC~\cite{kheradmand20243d} & \cellcolor{orange!25}29.30 & \cellcolor{orange!25}0.8715 & \cellcolor{orange!25}0.3282 & 7 MB & 786 ± 84 & 30K & \cellcolor{orange!25}24.02 & \cellcolor{orange!25}0.8288 & \cellcolor{orange!25}0.2669 & 19 MB & 1330 ± 154 & 80K & \cellcolor{red!25}27.66 & \cellcolor{red!25}0.8227 & \cellcolor{orange!25}0.2723 & 38 MB & 942 ± 158 & 160K \\ 
Mini-Splatting~\cite{fang2024mini} & \cellcolor{yellow!25}28.00 & \cellcolor{yellow!25}0.8546 & \cellcolor{yellow!25}0.3469 & 7 MB & 503 ± 52 & 30K & \cellcolor{yellow!25}23.25 & \cellcolor{yellow!25}0.7995 & \cellcolor{yellow!25}0.2949 & 19 MB & 1280 ± 111 & 80K & \cellcolor{yellow!25}27.35 & \cellcolor{yellow!25}0.8032 & \cellcolor{yellow!25}0.2868 & 38 MB & 829 ± 106 & 160K \\ \hline 
2DGS~\cite{huang20242d} & 26.09 & 0.8369 & 0.3530 & 7 MB & 136 ± 14 & 30K & 17.22 & 0.6669 & 0.4446 & 18 MB & 224 ± 104 & 80K & 23.85 & 0.7444 & 0.3578 & 37 MB & 106 ± 28 & 160K \\ 
2DGS\textdagger~\cite{huang20242d} & 27.18 & 0.8520 & 0.3438 & 7 MB & 148 ± 18 & 30K & 20.82 & 0.7762 & 0.3223 & 18 MB & 240 ± 58 & 80K & 25.66 & 0.7778 & 0.3192 & 37 MB & 115 ± 23 & 160K \\


BBSplat(\textbf{Ours}) & \cellcolor{red!25}29.39 & \cellcolor{red!25}0.8830 & \cellcolor{red!25}0.2770 & 7 MB & 162 ± 18 & 15K & \cellcolor{red!25}24.32 & \cellcolor{red!25}0.8476 & \cellcolor{red!25}0.2143 & 19 MB & 230 ± 24 & 30K & \cellcolor{orange!25}27.49 & \cellcolor{orange!25}0.8175 & \cellcolor{red!25}0.2336  & 38 MB & 96 ± 11 & 60K \\ 

\end{tabular}
\end{adjustbox}

\caption{\textbf{Quantitative results with a fixed storage space.} We report metrics for the following datasets: Tanks\&Temples \cite{Knapitsch2017}, Mip-NeRF-360 \cite{barron2022mipnerf360}, and DTU \cite{jensen2014large}. We demonstrate metrics for a scenario where we limit the number of primitives to achieve a small storage space. \textdagger - denotes the method's version without using depth-normal regularization.}
\label{tab:metrics_fixed_N}
\end{table*}

\begin{table*}[tb]
\centering

\begin{adjustbox}{width=1\textwidth}
\setlength{\tabcolsep}{2pt}
\begin{tabular}{l|cccccc|cccccc|cccccc}
     & \multicolumn{6}{c|}{DTU}   & \multicolumn{6}{c|}{Tanks\&Temples} & \multicolumn{6}{c}{Mip-NeRF-360} \\ \hline
     & PSNR$\uparrow$  & SSIM$\uparrow$   & LPIPS$\downarrow$ & Storage$\downarrow$ & FPS$\uparrow$ & Points$\downarrow$ & PSNR$\uparrow$  & SSIM$\uparrow$   & LPIPS$\downarrow$  & Storage$\downarrow$ & FPS$\uparrow$ & Points$\downarrow$ & PSNR$\uparrow$  & SSIM$\uparrow$   & LPIPS$\downarrow$  & Storage$\downarrow$ & FPS$\uparrow$ & Points$\downarrow$ \\ \hline
3DGS~\cite{kerbl20233d} & 28.39 & \cellcolor{yellow!25}0.8775 & 0.2850 & 121MB & 364 ± 42 & 512K & \cellcolor{yellow!25}24.31 & 0.8628 &  0.1942 & 301 MB  & 441 ± 80 & 1274K & \cellcolor{yellow!25}28.97 & 0.8694 & \cellcolor{yellow!25}0.1845 & 751 MB & 299 ± 49 & 3174K \\ 
Mip-Splatting~\cite{Yu2023MipSplatting} & 28.30 & 0.8759 & 0.2690 & 176 MB & 219 ± 35 & 734K & \cellcolor{yellow!25}24.31 & \cellcolor{orange!25}0.8721 & \cellcolor{yellow!25}0.1765 & 405 MB  & 324 ± 60 & 1683K & \cellcolor{orange!25}29.19 & \cellcolor{orange!25}0.8795 & \cellcolor{orange!25}0.1654 & 950MB & 203 ± 37 & 3952K \\ 
3DGS-MCMC~\cite{kheradmand20243d} & \cellcolor{orange!25}29.68 & \cellcolor{red!25}0.8903 & \cellcolor{red!25}0.2544 & 121 MB & 214 ± 26 & 512K & \cellcolor{red!25}25.56 & \cellcolor{red!25}0.8816 & \cellcolor{red!25}0.1700 & 301 MB & 348 ± 63 & 1274K & \cellcolor{red!25}29.52 & \cellcolor{red!25}0.8858 & \cellcolor{red!25}0.1626 & 751 MB & 189 ± 48 & 3174K \\ 
Mini-Splatting~\cite{fang2024mini} & 26.59 & 0.8670 & \cellcolor{orange!25}0.2579 & 101 MB & 501 ± 27 & 427K & 24.27 & 0.8642 & 0.1926 & \cellcolor{red!25}59 MB & 1550 ± 226 & 249K & 28.78 & \cellcolor{yellow!25}0.8716 & 0.1891 & \cellcolor{red!25}112 MB & 918 ± 142 & 473K \\ \hline

2DGS~\cite{huang20242d} & 28.04 & 0.8726 & 0.2984 & \cellcolor{yellow!25}59 MB & 95 ± 7 & 252K & 21.06 & 0.7600 & 0.3318 & \cellcolor{orange!25}153 MB & 159 ± 51 & 659K & 28.15 & 0.8537 & 0.2144 & 530 MB & 51 ± 7 & 2278K \\ 
2DGS\textdagger~\cite{huang20242d} & 28.62 & 0.8765 & 0.2976 & \cellcolor{orange!25}54 MB & 94 ± 7 & 225K & 24.20 & 0.8570 & 0.2081 & 187 MB & 98 ± 16 & 803K & 28.70 & 0.8604 & 0.1995 & \cellcolor{yellow!25}520 MB & 55 ± 7 & 2234K \\ 

BBSplat(\textbf{Ours}) & \cellcolor{red!25}29.72 & \cellcolor{orange!25}0.8865 & \cellcolor{yellow!25}0.2609 & \cellcolor{red!25}32 MB &  92 ± 10 & 60K & \cellcolor{orange!25} 25.12 & \cellcolor{yellow!25}0.8683 & \cellcolor{orange!25}0.1728 & \cellcolor{yellow!25}169 MB & 47 ± 4 & 300K & 28.33 & 0.8450 & 0.1971& \cellcolor{orange!25}239 MB & 28 ± 2 & 400K \end{tabular}
\end{adjustbox}

\caption{\textbf{Quantitative results with a large number of primitives.} We report metrics for the following datasets: Tanks\&Temples \cite{Knapitsch2017}, Mip-NeRF-360 \cite{barron2022mipnerf360}, and DTU \cite{jensen2014large}. We provide metrics for the default number of Gaussians for each method, which results in increased storage space.\textdagger - indicates method version without using depth-normal regularizations.}
\label{tab:metrics}
\end{table*}

\begin{figure*}[t]
  \centering
  \includegraphics[width=17.5cm]{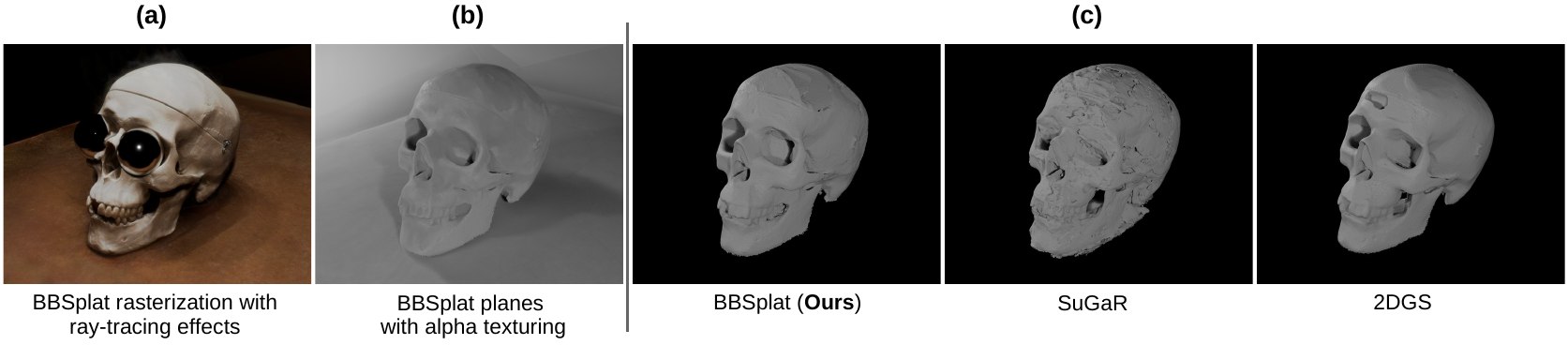}

   \caption{\textbf{Ray-tracing and mesh extraction.} a) BBSplat representation of the scene as explicit textured planes allows the implementation of ray-tracing effects during rasterization. b) Rasterization of planes with alpha texture provide accurate objects suitable for mesh extraction. c) Alternatively, our representation can be used to extract accurate 3D meshes, surpassing the accuracy of 3D Gaussians-based methods (\eg SuGaR). More renderings and extracted meshes can be found in the \cref{sec:results_images} of Supp. Mat..} 
   \label{fig:blender_mesh}
\end{figure*}

\begin{table}[]
\vspace{-0.5em}
\fontsize{7.5}{10}\selectfont
\centering
\begin{tabular}{c|cccc}
           & PSNR$\uparrow$  & SSIM$\uparrow$   & LPIPS$\downarrow$ & Storage$\downarrow$ \\ \hline
3DGS~\cite{kerbl20233d} & 28.39 & 0.8775 & 0.2850 & 121MB \\
Reduced-3DGS~\cite{papantonakis2024reducing} & 28.31 & 0.8769 & 0.2859 & 30 MB \\ 

Compact-3DGS~\cite{lee2024c3dgs} & 28.39 & 0.8752 & 0.3000 & 10MB \\

BBSplat(\textbf{Ours}) & \cellcolor{red!25}29.39 & \cellcolor{red!25}0.8830 & \cellcolor{red!25}0.2770 & \cellcolor{red!25}7 MB

\end{tabular}
\caption{\textbf{Comparison with compression methods on DTU.} We provide better NVS than state-of-the-art compression methods while requiring less storage space.}
\label{tab:compression}
\vspace{-1em}
\end{table}

\subsection{Implementation}
\label{sec:implement}
We use $\lambda_\textrm{SSIM}=0.2$ for the structure similarity in the \Cref{eq:photometric}, and $\lambda_\textrm{RGB}=\lambda_\alpha=1\mathrm{e}{-4}$ for the \Cref{eq:texture_reg}.
We initialize $T_\textrm{RGB}$ and $T_\alpha$ with zeros and 2D Gaussian transparency respectively, and keep them fixed for the first 500 iterations to adjust orientations and colors initialized with SfM. 
The densification strategy described in \cref{sec:BBSplat_MCMC} is used from 500 until 25,000 iterations to adjust the number of billboards. Overall, we train BBSplat for 30,000 steps and finally, we fine-tune the spherical harmonics $\textrm{SH}_i$ for an additional 2,000 iterations to adjust them for textures.

For training, we use the Adam \cite{kingma2014adam} optimizer with hyper-parameters recommended for the Gaussian splatting pipelines \cite{kerbl20233d}. We set RGB and $\alpha$ textures size $S_T=16$. resulting in $16\times16$ textures to get the best trade-off between quality and memory requirements. To optimize RGB-texture we set learning rate $lr_\textrm{RGB} = 2.5\mathrm{e}{-3}$ and for $\alpha$-texture we set $lr_\alpha=1\mathrm{e}{-3}$. Additionally, we found it beneficial for convergence to increase spherical harmonics learning rate to $lr_\textrm{SH}=5\mathrm{e}{-3}$. For more details, please refer to the supplementary materials. 


\subsection{Results and Evaluation}
\label{seq:results}
We compared with previous state-of-the-art methods on real-world scenes using standard datasets. We evaluated our method on 7 publicly available scenes from the Mip-NeRF-360 dataset with indoor and outdoor real scenes \cite{barron2022mipnerf360} and also on 5 outdoor scenes from the Tanks\&Temples dataset \cite{Knapitsch2017}. Furthermore, we evaluated our method on 15 scenes from DTU dataset \cite{jensen2014large}. Overall, we provide experiments for 27 real scenes from three datasets. 

We used train/test split for all datasets following 3DGS methodology where each 8$^{th}$ image was taken for the test. 
For all three datasets, we use the same image resolution as previous methods for fair comparison (\eg~3DGS). Namely, we reduced the original images from Tanks\&Temples $\times2$ times (to $\sim$900px width), Mip-NeRF-360 indoor scenes $\times2$ times (to $\sim$1600px width), outdoor scenes $\times4$ times (to $\sim$1200px width). For DTU we used original image size ($1600\times1200$px).
All experiments are conducted on a single NVIDIA GeForce RTX 4090 GPU.

\textbf{Quantitative comparison.}
We report PSNR, SSIM, and LPIPS \cite{zhang2018unreasonable} metrics (\cref{tab:metrics_fixed_N,tab:metrics}) widely used to evaluate quality in the NVS task. PSNR and SSIM are traditionally used to demonstrate similarity with ground truth images; LPIPS utilizes neural networks and it is known for the best correlation with human perception. 
In the provided experiments, we evaluated metrics for limited storage space constraining the amount of primitives by limiting the growth of their number during training as proposed in \cite{bulo2024revising}.

In an experiment in \Cref{tab:metrics_fixed_N}, we showcase metrics for the fixed storage space for different NVS methods. For the same storage space, our method reaches better PSNR, SSIM and LPIPS than state-of-the-art Gaussian-based representations both for indoor DTU and outdoor Tanks\&Temples scenes. For the Mip-NeRF-360 dataset, BBSplat reaches the best LPIPS even for highly detailed outdoor scenes challenging for textured representation. Notice also that BBSplat at fixed memory uses less primitives than competing methods, showing the representation power of textured Gaussians. 

In \Cref{tab:metrics}, we compare against other methods for the maximum recommended amount of Gaussians to achieve the best possible quality for each method. BBSplat achieves state-of-the-art PSNR value for indoor DTU dataset while reaching up to $\times5$ storage space reduction. For Tanks\&Temples outdoor scenes, we provide second best NVS quality reducing storage almost $\times2$ times. The Mip-NeRF-360 dataset is particularly challenging for texture-based representation as it contains many highly detailed outdoor scenes. Even in such a disadvantageous scenario, BBSplat reaches a high NVS quality. In the Supp. Mat. we provide per-scene metrics demonstrating that BBSplat provides high metrics values for indoor Mip-NeRF-360 scenes and lower metrics for outdoor scenes, which is one of the method's limitations.

In \Cref{tab:compression}, we provide a comparison with state-of-the-art compression methods for 3DGS. All provided methods are trained from scratch and apply various techniques to reduce storage space. BBSplat achieved the highest storage space reduction compared to 3DGS ($\times17$ times) while providing state-of-the-art rendering quality. 

\textbf{Qualitative comparison.}
We provide qualitative results in \Cref{fig:results} for indoor and outdoor scenes from different datasets. We show novel views for basic 3DGS and 2DGS representations and 3DGS-MCMC as they are the best scoring competitors. While the quality of the central object of the scene remains similar for all methods, in most scenarios BBSplat represents background objects with more details. For Tanks\&Temples and Mip-NeRF-360 dataset, we reached quality on par with state-of-the-art.
For the DTU dataset, we also demonstrate better handling of challenging light effects. For more results, please refer to Supp. Mat.. 

\textbf{Mesh Extraction.}
In \Cref{tab:chamfer}, we compare the 3D mesh extraction performance of BBSPlat with state-of-the-art methods on the DTU dataset \cite{jensen2014large}. For the experiment, we use 15 scenes with ground truth obtained by 3D scanning objects in a laboratory setting. For mesh extraction, we reduced input resolution to $800\times600$ for compatibility with the values reported by 2DGS. As a metric, we present the Chamfer Distance (CD) indicating the similarity between two meshes. We also report PSNR for NVS with HD resolution and each 8$^{th}$ frame held for evaluation.

Both experiments for 2DGS and for BBSplat were conducted using the depth-normal regularization proposed in \cite{huang20242d}. Rendering quality metrics for BBSplat in Tables \ref{tab:metrics_fixed_N} and \ref{tab:metrics} were also obtained using such regularizations.
BBSplat reaches 0.91 CD on DTU dataset, which is better than 1.96 for 3DGS and 1.33 for SuGaR~\cite{guedon2024sugar}, and close to 0.80 for 2DGS. We provide extracted mesh visualization in \Cref{fig:blender_mesh} (c). At the same time, BBSplat provides the best novel view rendering quality (PSNR) and requires the least storage space.

\textbf{Ray-tracing rasterization.}
Representing the scene as textured billboards allows us to consider it as a set of planes forming a textured mesh. Thus, we can rasterize such a plane-based mesh using classical computer graphic rasterization, enabling ray-tracing effects as shown in \Cref{fig:blender_mesh} (a). The image was obtained by rasterizing in Blender a set of planes with RGB and alpha textures. This way it is not necessary to spend extra time for extracting mesh representation and it allows  to preserve photo realistic rendering while enabling rasterization effects and editing. For animated examples, please refer to the supplementary video.

To reduce the number of overlapping artifacts during rasterization, we used StopThePop \cite{radl2024stopthepop} per-ray sorting algorithm during training. Note that StopThePop in this work was used only for visualizations in \Cref{fig:blender_mesh}. After training, we convert each billboard to the set of two polygons forming a plane with corresponding UV coordinates on the concatenated texture map.

\begin{table}[tb]
\fontsize{7.5}{10}\selectfont
\centering
\begin{tabular}{l|lll}
Method        & CD$\downarrow$ & PSNR$\uparrow$ & Storage$\downarrow$\\ \hline
3DGS~\cite{kerbl20233d}          & 1.96              & \cellcolor{yellow!25}28.39 & \cellcolor{yellow!25}121 MB\\
SuGaR~\cite{guedon2024sugar}     & \cellcolor{yellow!25}1.33              & 27.95 & 1059 MB\\
2DGS~\cite{huang20242d}          & \cellcolor{red!25}0.80              & \cellcolor{orange!25}28.04 & \cellcolor{orange!25}59 MB\\
BBSplat(\textbf{Ours})           & \cellcolor{orange!25}0.91   & \cellcolor{red!25}29.72 & \cellcolor{red!25}32 MB\end{tabular}

\caption{\textbf{Chamfer distance on the DTU dataset \cite{jensen2014large}} We report the Chamfer distance on 15 scenes from DTU dataset to evaluate 3D mesh extraction quality of the method.}
\label{tab:chamfer}
\vspace{-1em}
\end{table}

\subsection{Ablation study}

\begin{table}[tb]
\centering
\fontsize{7.5}{10}\selectfont

\begin{tabular}{l|cccc}
 & PSNR$\uparrow$ & SSIM$\uparrow$ & LPIPS$\downarrow$ & Storage$\downarrow$ \\ \hline
w/o $T^\textrm{RGB}$ & 25.17 & 0.8684 & 0.1774 & \cellcolor{red!25}60 MB  \\
w/o $T^{\alpha}$ & 25.26 & 0.8710 & 0.1858 & \cellcolor{orange!25}72 MB  \\
w/o $\mathcal{L_\textrm{texture}}$ & \cellcolor{yellow!25}25.54 & \cellcolor{yellow!25}0.8791 & \cellcolor{yellow!25}0.1655 & 131 MB  \\
w/o compress & \cellcolor{red!25}25.71 & \cellcolor{red!25}0.8818 & \cellcolor{red!25}0.1630 &  662 MB \\ \hline
Full & \cellcolor{orange!25}25.68  & \cellcolor{orange!25}0.8797 & \cellcolor{orange!25}0.1649 & \cellcolor{yellow!25}93 MB  
\end{tabular}

\caption{\textbf{Ablation study on Tanks\&Temples dataset.} We report the quantitative metrics for our method without the use of color or transparency textures, and without texture regularization and compression post-processing.}
\label{tab:ablation}
\vspace{-1em}
\end{table}

In \Cref{tab:ablation}, we provide an ablation study of different aspects of our approach on Tanks\&Temples dataset with $800\times600$ resolution. 
The use of both alpha and RGB textures is essential for rendering quality. Excluding $T^\alpha$ significantly increases LPIPS, while excluding $T^\textrm{RGB}$ reduces PSNR and SSIM. Texture regularization term preserves over-fitting and leads to additional metrics improvement. It also encourages sparsity in textures and reduces storage costs in combination with compression. The use of compression leads to a slight metric reduction but allows to significantly reduce storage space requirements.
Therefore, all parts of the proposed approach are essential to get the best result. However, disabling compression could further improve metrics. 

\section{Conclusion}
\label{sec:conclusion}
We have proposed BBSplat - a novel method for 3D scene representation. The proposed method uses textured primitives of arbitrary learnable shapes and allows accurate mesh extraction and photorealistic novel views synthesis of real scenes.
We developed specialized regularization term and compression technique to reduce storage space by leveraging the sparse nature of the proposed textured representation. In extensive experiments on real data, we demonstrated the efficiency of the proposed method both quantitatively and qualitatively. 

\noindent \paragraph{Acknowledgements.} We thank Matteo Toso, Milind Gajanan Padalkar, Matteo Bortolon, and Fabio Poiesi for their feedback and helpful suggestions on improving the paper. We also thank Andrea Tagliasacchi for insightful discussions and Matteo Bortolon for the help with CUDA implementation. D. Svitov thanks the ELLIS network for organizing the PhD program that made this work possible.

{
    \small
    \bibliographystyle{ieeenat_fullname}
    \bibliography{main}
}


\clearpage
\setcounter{page}{1}
\maketitlesupplementary

\begin{table}[]
\fontsize{8.5}{11}\selectfont
\centering
\begin{tabular}{l|ccccc}
      & 10 & 100 & \textbf{500} & 1000 & 1500 \\ \hline
PSNR$\uparrow$  &  \cellcolor{yellow!25}25.95  & \cellcolor{orange!25}25.96 & \cellcolor{red!25}25.99 & 25.92 & 25.92 \\
SSIM$\uparrow$  & \cellcolor{red!25}0.8800 & 0.8798 & \cellcolor{orange!25}0.8798 & 0.8798 & \cellcolor{yellow!25}0.8795 \\
LPIPS$\downarrow$ & \cellcolor{red!25}0.1365 & \cellcolor{orange!25}0.1375 & \cellcolor{yellow!25}0.1379 & 0.1392 & 0.1396 \\ \hline
Size(MB)$\downarrow$ & 139 & 122 &  \cellcolor{yellow!25}102 & \cellcolor{orange!25}92 & \cellcolor{red!25}87 
\end{tabular}
\caption{\textbf{Selecting $\sigma$ threshold value.} We search for the best $\sigma$ threshold in \Cref{eq:texture_reg} using ``Truck'' scene from the Tanks\&Temples dataset.}
\label{tab:threshold}
\end{table}

\begin{table}[]
\fontsize{7.5}{10}\selectfont
\centering
\begin{tabular}{l|cccc}
      & $8^2\times640K$ & $\boldsymbol{16^2\times160K}$ & $32^2\times40K$ & $64^2\times10K$ \\ \hline
PSNR$\uparrow$  & \cellcolor{red!25}26.04 & \cellcolor{orange!25}25.99 & \cellcolor{yellow!25}25.28 & 23.34 \\
SSIM$\uparrow$  & \cellcolor{red!25}0.8807 & \cellcolor{orange!25}0.8798 & \cellcolor{yellow!25}0.8662 & 0.8000 \\
LPIPS$\downarrow$ & \cellcolor{red!25}0.1393 & \cellcolor{orange!25}0.1379 & \cellcolor{yellow!25}0.1505 & 0.2382 \\ \hline
Size(MB)$\downarrow$ & 207 & \cellcolor{yellow!25}102 & \cellcolor{orange!25}82 & \cellcolor{red!25}68 
\end{tabular}
\caption{\textbf{Selecting texture size and number of primitives.} We search for best texture size and primitives number for the fixed parameters amount (4 channels x 40.96M). We evaluate metrics using the ``Truck'' scene from the Tanks\&Temples dataset.}
\label{tab:sizes}
\end{table}

\section{Implementation details}
\label{sec:details}

In this section, we provide more details on the method's implementation. In the following subsections, we describe the implementation of backpropagation for the bilinear texture sampling with billboard position adjustment and discuss the choice of hyper-parameters to train billboard splatting.

\subsection{Texture sampling}
We leverage \textit{bilinear sampling} to get texture values from $T^\textrm{RGB}_i$ and $T^\alpha_i$ in the $\textbf{u}=(u, v)$ point of a billboard. Our implementation utilizes GPU programmed with CUDA to accelerate sampling and loss backpropagation to the texture. We based our implementation on the \texttt{PyTorch}\cite{NEURIPS2019_9015} CUDA code for bilinear sampling and integrated it into 2DGS CUDA implementation as described below. 

In \Cref{alg:bilinear}, we provide pseudo-code for the bilinear sampling gradients backpropagation. The algorithm accepts the gradient value that should be distributed among neighboring texels, $(u, v)$ coordinate of the billboard point where the gradient is calculated, corresponding source texture, and output tensor to store texture gradients. The function redistributes and stores weighted values of the gradient in four neighboring texels and calculate the gradient $dL / d\textbf{u}$ with respect to point $(u, v)$ position.

In \Cref{alg:optimize}, we demonstrate how we prepare input parameters for \Cref{alg:bilinear} to get $dL / d\textbf{u}$ values based on RGB texture and alpha map. Calculated $dL / d\textbf{u}$ value is then used to adjust billboard position and rotation. The proposed implementation can be easily integrated with the 2DGS approach, which we encapsulate here as subprogram $\mathcal{F}$. The function $\mathcal{F}$ encapsulates gradients calculation for transformation matrix $\Sigma$ encoding translation and rotation of the plane. 

\begin{figure}[t]
  \centering
  \includegraphics[width=0.9\linewidth]{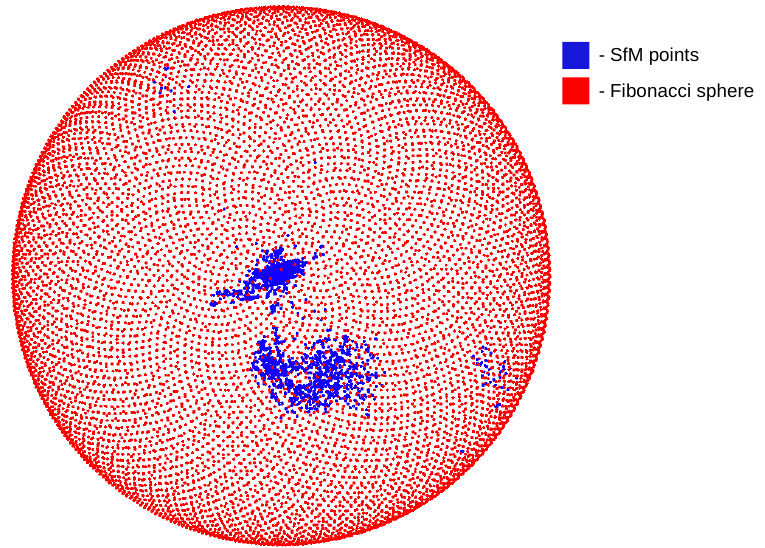}

   \caption{\textbf{Sky-sphere around the scene.} We sample additional points on the sphere around all SfM points with the Fibonacci algorithm \cite{gonzalez2010measurement}.}
   \label{fig:sphere}
\end{figure}

\begin{algorithm}
\caption{Function for bilinear gradient redistribution}\label{alg:bilinear}
\begin{algorithmic}

\Function{bilinear}{$\frac{dL}{d\omega}$, $(u, v)$, $T$, $\frac{dL}{dT}$} 

 \vspace{0.5em}
 \State $x = ((u + 1) / 2) (T_S - 1)$
 \State $y = ((v + 1) / 2) (T_S - 1)$

 \vspace{0.5em}
 \State $ne_x = \floor{x} + 1$ \quad  $se_x = \floor{x} + 1$  \quad $sw_x = \floor{x}$
 \State $ne_y = \floor{y}$ \quad $se_y = \floor{y} + 1$ \quad $sw_y = \floor{y} + 1$ 

 \vspace{0.5em}
 \State $nw = (se_x - x)(se_y - y)$ \quad $ne = (x - sw_x)(sw_y - y)$
 \State $sw = (ne_x - x)(y - ne_y)$ \quad $se = (x - nw_x)(y - nw_y)$

 \vspace{-0.5em}
  \texttt{\\Redistribute gradients}
 \State $\frac{dL}{dT}[nw_x, nw_y] = \frac{dL}{d\omega} \cdot nw$
 \State $\frac{dL}{dT}[ne_x, ne_y] = \frac{dL}{d\omega} \cdot ne$
 \State $\frac{dL}{dT}[sw_x, sw_y] = \frac{dL}{d\omega} \cdot sw$
 \State $\frac{dL}{dT}[se_x, se_y] = \frac{dL}{d\omega} \cdot se$

 \vspace{-0.5em}
  \texttt{\\Calculate position gradient}
 \State $g_x = -T[nw_x, nw_y] (se_y - y) \frac{dL}{d\omega}$
 \State $g_y = -T[nw_x, nw_y] (se_y - x) \frac{dL}{d\omega}$

 \State $g_x \mathrel{+}= T[ne_x, ne_y] (sw_y - y) \frac{dL}{d\omega}$
 \State $g_y \mathrel{-}= T[ne_x, ne_y] (x - sw_x) \frac{dL}{d\omega}$

 \State $g_x \mathrel{-}= T[sw_x, sw_y] (y - ne_y) \frac{dL}{d\omega}$
 \State $g_y \mathrel{+}= T[sw_x, sw_y] (ne_x - x) \frac{dL}{d\omega}$

 \State $g_x \mathrel{+}= T[se_x, se_y] (y - nw_y) \frac{dL}{d\omega}$
 \State $g_y \mathrel{+}= T[se_x, se_y] (x - nw_x) \frac{dL}{d\omega}$

 \vspace{1em}
 \State $g_x = g_x (T_s - 1) / 2$ \quad $g_y = g_y (T_s - 1) / 2$

 \vspace{1em}
 \Return $(g_x, g_y)$  \Comment{Values for $\frac{dL}{du}$}

\EndFunction

\end{algorithmic}
\end{algorithm}

\begin{algorithm}
\caption{Billboards position optimization}\label{alg:optimize}
\begin{algorithmic}
\Require $\frac{dL}{dI}$ \Comment{Gradient of loss $L$ w.r.t image $I$}
\Require $\textbf{u}$ \Comment{Plane intersection $(u, v)$ point}
\Require $\mathcal{B}$ \Comment{Billboards sorted along the ray}
\Require $\mathcal{T}$ \Comment{Forward pass accumulated transparency}

\vspace{0.5em}
\For{$i \in \mathcal{B}$}

  \vspace{-0.5em}
  \texttt{\\Sample alpha and color}
  \State $\alpha = T_i^\alpha[\textbf{u}]$
  \State $c = T_i^\textrm{RGB}[\textbf{u}] + \textrm{SH}_i$ 
  \State $\hat{c} = \alpha' c' + (1 - \alpha') c$ \Comment{Accumulated color}

  \vspace{-0.5em}
  \texttt{\\Update transparency}
  \State $\mathcal{T} = \mathcal{T} / (1 - \alpha)$

  \vspace{-0.5em}
  \texttt{\\Calculate gradients}
  \State $\frac{dL}{dc} = \alpha \mathcal{T} \frac{dL}{dI}$
  \State $\frac{dL}{d\alpha} = (c - \hat{c}) \mathcal{T} \frac{dL}{dI}$

  \vspace{-0.5em}
  \texttt{\\Redistribute gradients}
  \State $\frac{dL}{du} = \textrm{BILINEAR}(\frac{dL}{dc}, \textbf{u}, T_i^\textrm{RGB}, \frac{dL}{dT_i^\textrm{RGB}})$
  \State $\frac{dL}{du} \mathrel{+}= \textrm{BILINEAR}(\frac{dL}{d\alpha}, \textbf{u}, T_i^\alpha, \frac{dL}{dT_i^\alpha})$

  \vspace{-0.5em}
  \texttt{\\Adjust position}
  \State $\frac{dL}{d\Sigma} = \mathcal{F}(\frac{dL}{du})$  \Comment{Calculate $\frac{dL}{d\Sigma}$ as in 2DGS}

  \vspace{1em}
  \State $\alpha' = \alpha$
  \State $c' = c$
  
\EndFor
\end{algorithmic}
\end{algorithm}

\subsection{Hyper-parameters}
\label{sec:details_params}
We apply learning rate exponential decay for billboards position from $lr_\mu=1.6\mathrm{e}{-4}$ to $lr_\mu=1.6\mathrm{e}{-6}$ with $0.01$ multiplier. To optimize spherical harmonics and scales we set $lr_\textrm{SH}=ls_s=5\mathrm{e}{-3}$ while using smaller learning rate for rotation $le_r=1\mathrm{e}{-3}$. 

We select $\sigma$ threshold value in \Cref{eq:texture_reg} based on the metrics provided in \Cref{tab:threshold}. The $\sigma$ threshold is used to determine the minimum billboard visibility from which we start regularizing its textures. We set $\sigma=500$ as it provides the best PSNR value and achieves the best trade-off between other metrics and storage space. Over-regularization results in more billboards having Gaussian distributed transparency, which leads to more compact storing of them but reduces NVS quality. Under-regularization allows to achieve  higher SSIM and LPIPS at the cost of storage space and inference speed. 

\Cref{tab:sizes} shows metrics for different texture sizes and number of billboards. We select texture size to be the power of two (8, 16, 32, 64) as it is more suitable for processing with GPU, and set the corresponding number of billboards to result in 40.96M parameters for one texture channel to not exceed GPU capacity. While $32 \times 32$ and $16 \times 16$ both provide suitable metrics, we chose $16 \times 16$ as slightly more accurate. 

\subsection{Initialization} 
In scenarios when we need fewer SfM points to initialize a small number of billboards, we subsample them with an iterative farthest point sampling strategy. For outdoor scenes, we optionally add 10K evenly distributed points on the large sphere using the Fibonacci algorithm \cite{gonzalez2010measurement} to represent the sky and far away objects. We select sphere radius as the distance from the scene center to the furthest SfM point to guarantee that all SfM points are included in the sphere. For scenarios when we use 20K points and less, we reduce the number of sphere points to 2K, and completely disable it when we use less than 10K points. In \Cref{fig:sphere}, we provide a visualization of such a sphere.

\section{Additional details on experiments and results}
\label{sec:results}

In this section, we provide more qualitative results for our method along with per-scene quantitative metrics for averaged values reported in the paper.

\subsection{Visual results}
\label{sec:results_images}
\begin{figure*}
    \centering
    \includegraphics[width=17.5cm]{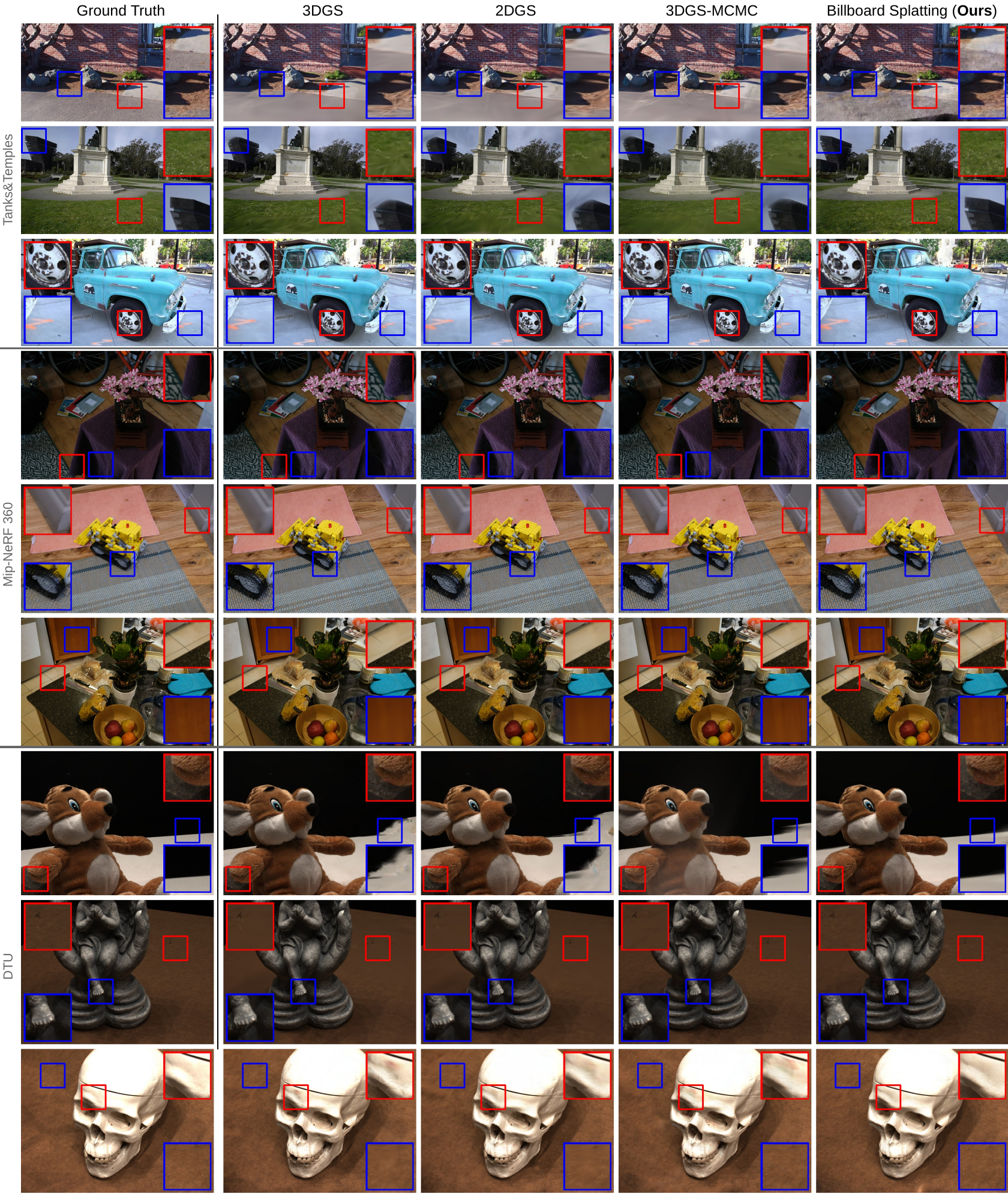}
    \caption{\textbf{Qualitative results.} We provide rendering results of three more scenes from each dataset: Tanks\&Temples [Lighthouse, Francis, Truck], Mip-NeRF-360 [Bansai, Kitchen, Counter], and DTU [Scan105, Scan118, Scan65]. For competitors, we use the maximum number of Gaussians recommended by the method. 
    }
    \label{fig:results_supmat}
\end{figure*}

In \Cref{fig:results_supmat}, we provide an additional visual comparison of BBSplat with state-of-the-art NVS methods. Our renderings demonstrate exceeding quality. 

\begin{figure*}[t]
  \centering
  \includegraphics[width=17.5cm]{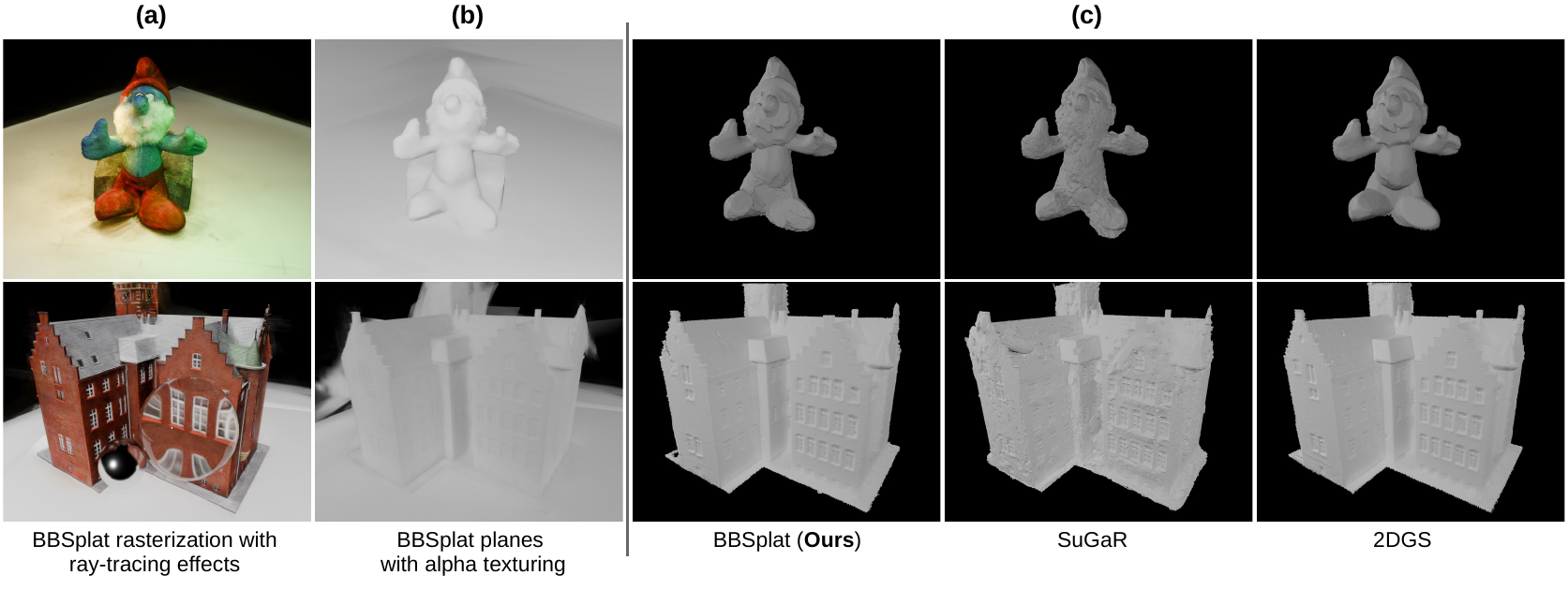}

   \caption{\textbf{Ray-tracing and mesh extraction.} a) BBSplat representation of the scene as explicit textured planes allows the implementation of ray-tracing effects during rasterization (\eg religtning or mirror effects). BBSplat representation can be rasterized as splatting or with classic rasterization tools (\eg Blender). b) Rasterization of planes with alpha texture provide accurate objects suitable for mesh extraction. c) Alternatively, our representation can be used to extract accurate 3D meshes, surpassing the accuracy of 3D Gaussians-based methods (\eg SuGaR).}
   \label{fig:blender_mesh_suppmat}
\end{figure*}

In \Cref{fig:blender_mesh_suppmat}, we demonstrate additional renderings with ray-tracing effects and extracted meshes for the DTU dataset.

\begin{figure*}[t]
  \centering
  \includegraphics[width=0.95\linewidth]{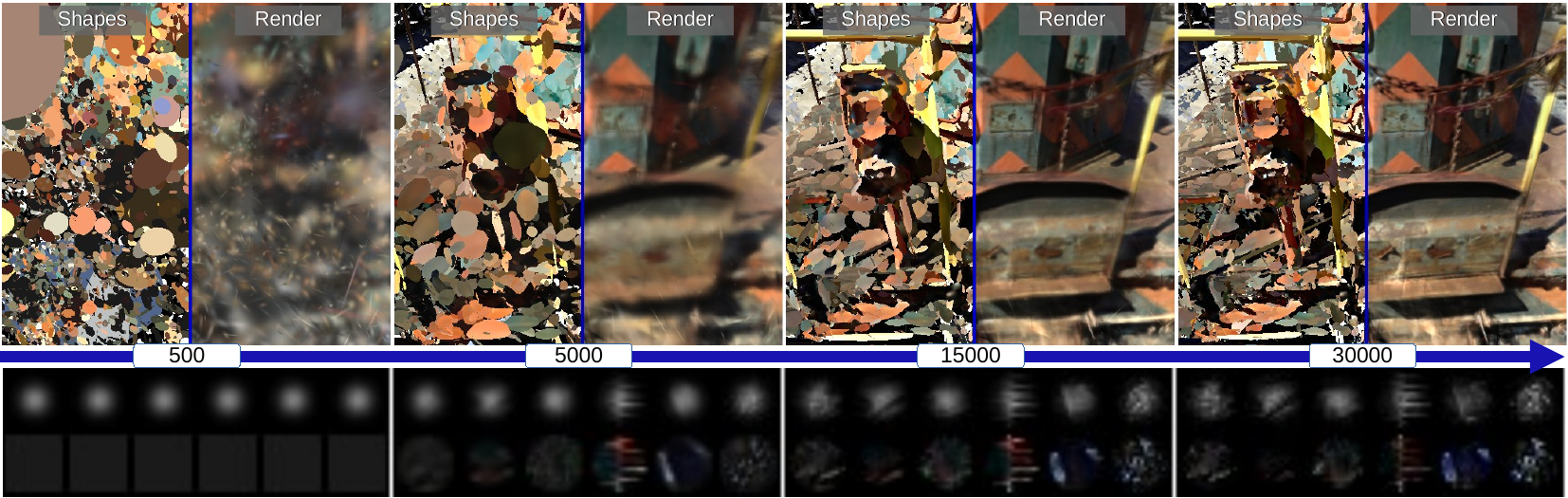}

   \caption{\textbf{BBsplat optimization process}. Top: We demonstrate billboard silhouettes and rendering results during the training process for 500, 5000, 15000, and 30000 iterations. To visualize silhouettes we disabled color textures and cut alpha-maps to the threshold. Bottom: We show examples of alpha-maps and RGB textures corresponding to training steps.}
   \label{fig:progress}
   
\vspace{-0.3em}
\end{figure*}

\Cref{fig:progress} demonstrates BBSplat training progress for 500, 5000, 15000, and 30000 iterations, showing the change of transparency from Gaussian distribution to arbitrary shapes.

\subsection{Detailed results}
In \Cref{tab:scene_results}, we report scene-by-scene results on the experiments provided in \Cref{tab:metrics_fixed_N}.  We demonstrate state-of-the-art metrics quality for the Mip-NeRF-360 dataset on the indoor scenes while falling shortly behind for the outdoor scenes. This is caused by a huge number of small monochromic details (\eg leafs) that are more efficient to represent with Gaussian primitives. For the Tanks\&Temples outdoor scenes and DTU indoor scenes, we provide the best NVS for all cases. 

\subsection{Comparison to Texture-GS}
In \Cref{tab:texture_gs}, we compare with Texture-GS \cite{xu2024texture} as it also proposes to employ textures, although in conjunction with a 3DGS model.
We report metrics only for the DTU dataset since Texture-GS is unable to process large scenes, as it is limited by spherical texture space definition.

\begin{table}[]
\vspace{-0.5em}
\fontsize{7.5}{10}\selectfont
\centering
\begin{tabular}{c|cccc}
           & PSNR$\uparrow$  & L1$\downarrow$ & LPIPS$\downarrow$ \\ \hline
Texture-GS & 30.03 & 0.0135 & 0.1440 \\
BBSplat(\textbf{Ours}) & \cellcolor{red!25}30.81 & \cellcolor{red!25}0.0123 & \cellcolor{red!25}0.1055
\end{tabular}
\caption{\textbf{Comparison with Texture-GS \cite{xu2024texture} on DTU.} While Texture-GS also applies textures for Gaussians, it does not perform well for NVS.}
\label{tab:texture_gs}
\vspace{-1em}
\end{table}

\subsection{Limitations and future work}
One of the  limitations of the proposed method is the NVS for the large outdoor scenes. While we provide high rendering quality for Tanks\&Temples, there is still room for improvement in future works for the Mip-NeRF-360 dataset. Another limitation is training time, as it takes about 40 minutes to fit a scene, compared with 5 minutes for 3DGS. This slowdown is caused by backpropagation to the textures, and while it still significantly outperforms NeRF-based methods in terms of speed, it could be a bottleneck for some applications. Another limitation is that the use of a higher amount of primitives for outdoor scenes leads to over-fitting. In future works, we are going to focus on resolving these limitations.

\begin{table*}[]
\fontsize{8}{12}\selectfont
\centering
\begin{tabular}{cc|cccc}
                                &            & 3DGS~\cite{kerbl20233d}    & 3DGS-MCMC~\cite{kheradmand20243d}       & 2DGS\textdagger~\cite{huang20242d}  & BBSplat (\textbf{Ours})  \\ 
                                &            & \footnotesize{PSNR$\uparrow$ / SSIM$\uparrow$ / LPIPS$\downarrow$} & \footnotesize{PSNR$\uparrow$ / SSIM$\uparrow$ / LPIPS$\downarrow$} & \footnotesize{PSNR$\uparrow$ / SSIM$\uparrow$ / LPIPS$\downarrow$} & \footnotesize{PSNR$\uparrow$ / SSIM$\uparrow$ / LPIPS$\downarrow$} \\ \hline
\parbox[t]{2mm}{\multirow{5}{*}{\rotatebox[origin=c]{90}{\footnotesize{Tanks\&Temples}}}} 
                                & Train      &  \colorbox{yellow!30}{19.81} / \colorbox{yellow!30}{0.7229} / \colorbox{yellow!30}{0.3347}&  \colorbox{orange!30}{21.21} / \colorbox{orange!30}{0.7522} / \colorbox{orange!30}{0.3120}& 19.35 / 0.7149 / 0.3417& \colorbox{red!30}{21.38} / \colorbox{red!30}{0.7702} / \colorbox{red!30}{0.2604}\\
                                & Truck      &  20.43 / \colorbox{yellow!30}{0.7511} / \colorbox{yellow!30}{0.3192}&  \colorbox{orange!30}{23.91} / \colorbox{orange!30}{0.8245} / \colorbox{orange!30}{0.2513}& \colorbox{yellow!30}{20.77} / 0.7479 / 0.3283& \colorbox{red!30}{24.19} / \colorbox{red!30}{0.8418} / \colorbox{red!30}{0.1957}\\
                                & Francis    &  \colorbox{yellow!30}{24.16} / \colorbox{yellow!30}{0.8456} / \colorbox{yellow!30}{0.3368}& \colorbox{orange!30}{27.82} / \colorbox{orange!30}{0.8813} / \colorbox{orange!30}{0.2964}& 22.61 / 0.8356 / 0.3485& \colorbox{red!30}{28.56} / \colorbox{red!30}{0.9079} / \colorbox{red!30}{0.2334}\\
                                & Horse      & 19.17 / \colorbox{yellow!30}{0.7831} / \colorbox{yellow!30}{0.2973}& \colorbox{red!30}{24.93} / \colorbox{orange!30}{0.8695} / \colorbox{orange!30}{0.2080}& \colorbox{yellow!30}{19.92} / 0.7775 / 0.3116& \colorbox{orange!30}{24.72} / \colorbox{red!30}{0.8847} / \colorbox{red!30}{0.1604}\\
                                & Lighthouse & 20.93 / 0.8044 / \colorbox{yellow!30}{0.2773}& \colorbox{orange!30}{22.25} / \colorbox{orange!30}{0.8163} / \colorbox{orange!30}{0.2668}& \colorbox{yellow!30}{21.45} / \colorbox{yellow!30}{0.8053} / 0.2817& \colorbox{red!30}{22.75} / \colorbox{red!30}{0.8332} / \colorbox{red!30}{0.2214}\\ \hline
\parbox[t]{2mm}{\multirow{7}{*}{\rotatebox[origin=c]{90}{\footnotesize{Mip-NeRF-360}}}}   
                                & Room       & 29.24 / \colorbox{yellow!30}{0.8924} / \colorbox{yellow!30}{0.2721}& \colorbox{orange!30}{30.93} / \colorbox{orange!30}{0.9056} / \colorbox{orange!30}{0.2541}& \colorbox{yellow!30}{29.76} / 0.8914 / 0.2731 & \colorbox{red!30}{31.13} / \colorbox{red!30}{0.9146} / \colorbox{red!30}{0.2150} \\
                                & Kitchen    & 24.56 / 0.8712 / 0.2129& \colorbox{orange!30}{29.16} / \colorbox{orange!30}{0.9003} / \colorbox{orange!30}{0.1787}& \colorbox{yellow!30}{25.93} / \colorbox{yellow!30}{0.8745} / \colorbox{yellow!30}{0.2095}& \colorbox{red!30}{30.35} / \colorbox{red!30}{0.9116} / \colorbox{red!30}{0.1430}\\
                                & Bonsai     & 25.47 / 0.8856 / 0.2986& \colorbox{orange!30}{30.48} / \colorbox{red!30}{0.9230} / \colorbox{orange!30}{0.2447}& \colorbox{yellow!30}{26.24} / \colorbox{yellow!30}{0.8895} / \colorbox{yellow!30}{0.2944}& \colorbox{red!30}{29.62} / \colorbox{orange!30}{0.9085} / \colorbox{red!30}{0.2334}\\
                                & Counter    & 26.36 / \colorbox{yellow!30}{0.8716} / \colorbox{yellow!30}{0.2628}& \colorbox{orange!30}{28.28} / \colorbox{orange!30}{0.8910} / \colorbox{orange!30}{0.2350}& \colorbox{yellow!30}{26.65} / 0.8701 / 0.2649& \colorbox{red!30}{28.30} / \colorbox{red!30}{0.8941} / \colorbox{red!30}{0.2061}\\ \cline{2-6} 
                                & Bicycle    & \colorbox{yellow!30}{23.04} / \colorbox{yellow!30}{0.5912} / \colorbox{yellow!30}{0.4354}& \colorbox{red!30}{23.87} / \colorbox{red!30}{0.6675} / \colorbox{orange!30}{0.3652}& 22.74 / 0.5738 / 0.4460& \colorbox{orange!30}{23.28} / \colorbox{orange!30}{0.6520} / \colorbox{red!30}{0.3096}\\ 
                                & Garden    & 23.98 / \colorbox{yellow!30}{0.7093} / \colorbox{yellow!30}{0.3400}& \colorbox{orange!30}{25.06} / \colorbox{orange!30}{0.7512} / \colorbox{orange!30}{0.2974}& \colorbox{yellow!30}{24.26} / 0.7053 / 0.3419& \colorbox{red!30}{25.75} / \colorbox{red!30}{0.7873} / \colorbox{red!30}{0.1907}\\ 
                                & Stump    & \colorbox{yellow!30}{24.08} / 0.6387 / \colorbox{yellow!30}{0.4035}& \colorbox{orange!30}{25.82} / \colorbox{red!30}{0.7201} / \colorbox{red!30}{0.3311}& 24.05 / \colorbox{yellow!30}{0.6399} / 0.4047& 24.00 / \colorbox{orange!30}{0.6547} / \colorbox{orange!30}{0.3376}\\ \hline
\parbox[t]{2mm}{\multirow{15}{*}{\rotatebox[origin=c]{90}{\footnotesize{DTU}}}}  
                                & Scan24  & \colorbox{yellow!30}{22.09} / \colorbox{yellow!30}{0.8450} / \colorbox{yellow!30}{0.2590}& \colorbox{orange!30}{27.19} / \colorbox{orange!30}{0.8745} / \colorbox{orange!30}{0.2319}& 21.83 / 0.8423 / 0.2681& \colorbox{red!30}{27.47} / \colorbox{red!30}{0.8850} / \colorbox{red!30}{0.1677}\\
                                & Scan37  & 20.01 / 0.8097 / 0.2942& \colorbox{orange!30}{24.18} / \colorbox{orange!30}{0.8449} / \colorbox{orange!30}{0.2673}& \colorbox{yellow!30}{20.83} / \colorbox{yellow!30}{0.8187} / \colorbox{yellow!30}{0.2881}& \colorbox{red!30}{24.93} / \colorbox{red!30}{0.8670} / \colorbox{red!30}{0.2127}\\
                                & Scan40  & 19.21 / 0.7676 / 0.3535& \colorbox{orange!30}{22.46} / \colorbox{orange!30}{0.8132} / \colorbox{orange!30}{0.3196}& \colorbox{yellow!30}{20.47} / \colorbox{yellow!30}{0.7717} / \colorbox{yellow!30}{0.3493}& \colorbox{red!30}{25.56} / \colorbox{red!30}{0.8709} / \colorbox{red!30}{0.2081}\\
                                & Scan55  & 25.25 / 0.7934 / \colorbox{yellow!30}{0.3831}& \colorbox{red!30}{27.88} / \colorbox{orange!30}{0.8221} / \colorbox{orange!30}{0.3775}& 25.67 / \colorbox{yellow!30}{0.7959} / \colorbox{yellow!30}{0.3857}& \colorbox{orange!30}{27.74} / \colorbox{red!30}{0.8511} / \colorbox{red!30}{0.2746}\\
                                & Scan63  & 23.91 / 0.9098 / 0.2156& \colorbox{orange!30}{31.08} / \colorbox{orange!30}{0.9411} / \colorbox{orange!30}{0.1792}& \colorbox{yellow!30}{30.11} / \colorbox{yellow!30}{0.9228} / \colorbox{yellow!30}{0.1963}& \colorbox{red!30}{30.32} / \colorbox{red!30}{0.9422} / \colorbox{red!30}{0.1652}\\
                                & Scan65  & 27.76 / 0.8476 / \colorbox{yellow!30}{0.3648}& \colorbox{orange!30}{29.41} / \colorbox{orange!30}{0.8585} / \colorbox{orange!30}{0.3605}& \colorbox{yellow!30}{29.32} / \colorbox{yellow!30}{0.8513} / 0.3688& \colorbox{red!30}{29.67} / \colorbox{red!30}{0.8754} / \colorbox{red!30}{0.2803}\\
                                & Scan69  & 26.20 / \colorbox{yellow!30}{0.8485} / \colorbox{yellow!30}{0.3545}& \colorbox{red!30}{26.97} / \colorbox{orange!30}{0.8592} / \colorbox{orange!30}{0.3483}& \colorbox{yellow!30}{26.34} / 0.8465 / 0.3576& \colorbox{orange!30}{26.71} / \colorbox{red!30}{0.8660} / \colorbox{red!30}{0.2845}\\
                                & Scan83  & 28.36 / \colorbox{yellow!30}{0.8629} / \colorbox{yellow!30}{0.3888}& \colorbox{red!30}{29.62} / \colorbox{orange!30}{0.8679} / \colorbox{orange!30}{0.3818}& \colorbox{yellow!30}{28.69} / 0.8590 / 0.3921& \colorbox{orange!30}{28.97} / \colorbox{red!30}{0.8685} / \colorbox{red!30}{0.3604}\\
                                & Scan97  & 22.90 / 0.8263 / 0.3762& \colorbox{red!30}{27.67} / \colorbox{red!30}{0.8588} / \colorbox{orange!30}{0.3478}& \colorbox{yellow!30}{24.45} / \colorbox{yellow!30}{0.8360} / \colorbox{yellow!30}{0.3673}& \colorbox{orange!30}{27.39} / \colorbox{orange!30}{0.8579} / \colorbox{red!30}{0.3268}\\
                                & Scan105 & \colorbox{yellow!30}{24.58} / \colorbox{yellow!30}{0.8429} / \colorbox{yellow!30}{0.3781}& \colorbox{red!30}{30.10} / \colorbox{orange!30}{0.8687} / \colorbox{orange!30}{0.3579}& 23.92 / 0.8362 / 0.3797& \colorbox{orange!30}{29.35} / \colorbox{red!30}{0.8765} / \colorbox{red!30}{0.3154}\\
                                & Scan106 & \colorbox{yellow!30}{30.46} / \colorbox{yellow!30}{0.8780} / 0.3632& \colorbox{red!30}{33.38} / \colorbox{orange!30}{0.8983} / \colorbox{orange!30}{0.3455}& 30.39 / 0.8723 / \colorbox{yellow!30}{0.3629}& \colorbox{orange!30}{32.93} / \colorbox{red!30}{0.9020} / \colorbox{red!30}{0.3133}\\
                                & Scan110 & 30.71 / 0.8779 / 0.3773& \colorbox{red!30}{31.55} / \colorbox{red!30}{0.8898} / \colorbox{orange!30}{0.3666}& \colorbox{yellow!30}{31.12} / \colorbox{orange!30}{0.8806} / \colorbox{yellow!30}{0.3771}& \colorbox{orange!30}{31.42} / \colorbox{yellow!30}{0.8798} / \colorbox{red!30}{0.3324}\\
                                & Scan114 & 28.58 / \colorbox{yellow!30}{0.8642} / \colorbox{yellow!30}{0.3540}& \colorbox{red!30}{29.82} / \colorbox{orange!30}{0.8807} / \colorbox{orange!30}{0.3462}& \colorbox{yellow!30}{28.65} / 0.8636 / 0.3547& \colorbox{orange!30}{29.72} / \colorbox{red!30}{0.8870} / \colorbox{red!30}{0.2976}\\
                                & Scan118 & 31.77 / 0.8856 / \colorbox{yellow!30}{0.3561}& \colorbox{orange!30}{34.06} / \colorbox{orange!30}{0.9010} / \colorbox{orange!30}{0.3482}& \colorbox{yellow!30}{32.49} / \colorbox{yellow!30}{0.8886} / 0.3572& \colorbox{red!30}{34.41} / \colorbox{red!30}{0.9090} / \colorbox{red!30}{0.3077}\\
                                & Scan122 & 29.83 / 0.8640 / 0.3603& \colorbox{orange!30}{34.19} / \colorbox{orange!30}{0.8949} / \colorbox{orange!30}{0.3459}& \colorbox{yellow!30}{33.52} / \colorbox{yellow!30}{0.8956} / \colorbox{yellow!30}{0.3521}& \colorbox{red!30}{34.41} / \colorbox{red!30}{0.9069} / \colorbox{red!30}{0.3092}\end{tabular}
\caption{\textbf{Quantitative metrics for all scenes.} We report PSNR, SSIM, and LPIPS for each scene in all three datasets: Tanks\&Temples \cite{Knapitsch2017}, Mip-NeRF-360 \cite{barron2022mipnerf360}, and DTU \cite{jensen2014large}. For comparison, we provide metrics for basic 3DGS~\cite{kerbl20233d} and 2DGS~\cite{huang20242d} methods, and for 3DGS-MCMC~\cite{kheradmand20243d} as the best-scoring competitor.}
\label{tab:scene_results}
\end{table*}

\end{document}